\documentclass[10pt,twocolumn,letterpaper]{article}

\usepackage{iccv}
\usepackage{times}
\usepackage{epsfig}
\usepackage{graphicx}
\usepackage{amsmath}
\usepackage{amssymb}

\usepackage{microtype}
\usepackage{todonotes}
\usepackage{tikz}
\usepackage{pgfplots}
\usepackage{xcolor}
\usepackage{float}
\usepackage{bm}
\usepackage{multirow}
\usepackage{subfloat}
\usepackage{adjustbox}
\usepackage{wrapfig}
\usepackage{colortbl}
\usepackage{enumitem}
\usepackage[normalem]{ulem}
\usepackage{rotating}
\usepackage[caption=false]{subfig}

\DeclareMathAlphabet{\mathsfit}{\encodingdefault}{\sfdefault}{m}{sl}
\SetMathAlphabet{\mathsfit}{bold}{\encodingdefault}{\sfdefault}{bx}{n}
\newcommand{\tens}[1]{\bm{\mathsfit{#1}}}
\def\abbrmethod{WSCL}

\definecolor{mygray}{gray}{0.9}  
\definecolor{mybluecolor}{RGB}{0, 168, 234}
\definecolor{myredcolor}{RGB}{254, 0, 19}
\definecolor{fullcolor}{HTML}{CEE5D0}
\definecolor{weakcolor}{HTML}{E0C097}

\usepackage[pagebackref=true,breaklinks=true,letterpaper=true,colorlinks,bookmarks=false]{hyperref}

\usepackage[capitalize]{cleveref}
\crefname{section}{Sec.}{Secs.}
\Crefname{section}{Section}{Sections}
\Crefname{table}{Table}{Tables}
\crefname{table}{Tab.}{Tabs.}

\iccvfinalcopy 

\def\iccvPaperID{5349} 

\ificcvfinal\fi

\begin{document}

\title{Towards Generic Image Manipulation Detection with \\ Weakly-Supervised Self-Consistency Learning}

\author{Yuanhao Zhai ~~~~
Tianyu Luan ~~~~
David Doermann ~~~~
Junsong Yuan \\
University at Buffalo \\
{\tt\small \{yzhai6, tianyulu, doermann, jsyuan\}@buffalo.edu}
}

\maketitle
\ificcvfinal\thispagestyle{empty}\fi
\newcommand{\TL}[1]{\textbf{\textcolor{blue}{[Tianyu: #1]}}}

\begin{abstract}
As advanced image manipulation techniques emerge, detecting the manipulation becomes increasingly important.
Despite the success of recent learning-based approaches for image manipulation detection, they typically require expensive pixel-level annotations to train, while exhibiting degraded performance when testing on images that are differently manipulated compared with training images.
To address these limitations, we propose weakly-supervised image manipulation detection, such that only \emph{binary image-level labels} (authentic or tampered with) are required for training purpose.
Such a weakly-supervised setting can leverage more training images and has the potential to adapt quickly to new manipulation techniques.
To improve the generalization ability, we propose weakly-supervised self-consistency learning (\abbrmethod{}) to leverage the weakly annotated images.
Specifically, two consistency properties are learned: multi-source consistency (MSC) and inter-patch consistency (IPC).
MSC exploits different content-agnostic information and enables cross-source learning via an online pseudo label generation and refinement process.
IPC performs global pair-wise patch-patch relationship reasoning to discover a complete region of manipulation.
Extensive experiments validate that our \abbrmethod{}, even though is weakly supervised, exhibits competitive performance compared with fully-supervised counterpart under both in-distribution and out-of-distribution evaluations, as well as reasonable manipulation localization ability.
\end{abstract}

\section{Introduction}
\label{sec:introduction}

With the advent of increasingly powerful image editing
techniques~\cite{portenier2018faceshop,li2020manigan,liu2020open,xia2021tedigan,patashnik2021styleclip,jiang2021language,yu2019free,yang2020deep,zeng2021sketchedit,jiang2021language,shi2021learning},
image manipulation has never been so convenient, and can be easily
accomplished using natural
language~\cite{shi2021learning,jiang2021language,shi2021learning} or
sketch~\cite{portenier2018faceshop,yu2019free,yang2020deep,zeng2021sketchedit} by general users.
Such advances allow malicious users to easily manipulate images,
creating fake news, promoting blackmail, and generating
Deepfakes~\cite{zampoglou2017large,tolosana2020deepfakes}.
Thus, detecting the authenticity of an image is crucial for media
forensics and credible information sharing.

\begin{figure}[t]
   \centering
   \subfloat[In-distribution (IND) manipulation detection result on the testing split of the CASIA dataset~\cite{dong2010casia,dong2013casia}.]{\includegraphics[width=0.475\linewidth]{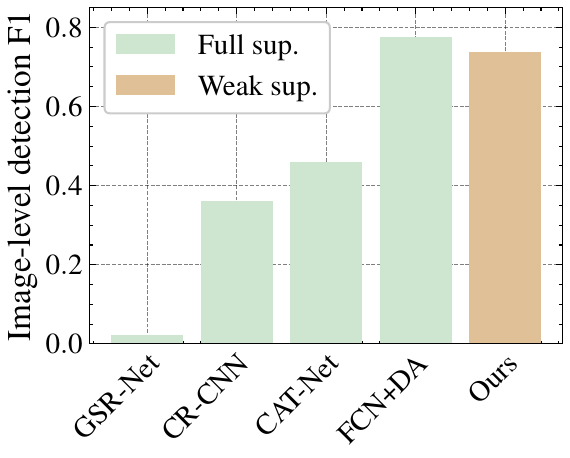}}
   \hfill
   \subfloat[Out-of-distribution (OOD) manipulation detection results, which are averaged over Columbia~\cite{hsu2006detecting}, Coverage~\cite{wen2016coverage} and IMD2020~\cite{novozamsky2020imd2020}.]{\includegraphics[width=0.475\linewidth]{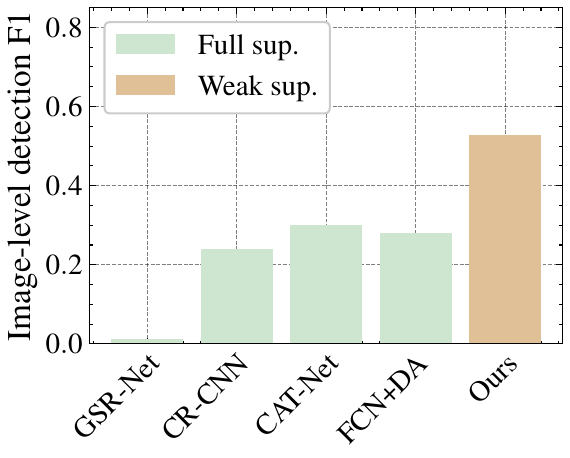}}
   \caption{Image-level manipulation detection performance comparison with
   existing fully-supervised
   methods~\cite{zhou2020generate,yang2020constrained,kwon2021cat,chen2021image}.
   All methods are trained on CASIA~\cite{dong2010casia,dong2013casia}.
   Our weakly-supervised method achieves comparable performance with
   fully-supervised methods under both IND and OOD manipulation detections.}
   \label{fig:teaser}
\end{figure}

Despite previous efforts to detect image manipulations, existing solutions still confront several challenges when dealing with real problems.
First, although learning-based image manipulation techniques demonstrate excellent performance compared with traditional methods, they may not easily generalize well to testing images that are manipulated differently compared with training images.
As more sophisticated image manipulation techniques continue to emerge, it is
exceedingly challenging, if not impracticable, to encompass all manipulation methods in the training data to enable effective handling of novel manipulations.
As shown in~\cref{fig:teaser}, despite work well in the training image dataset, the performance of learning-based methods can degrade considerably when testing on out-of-distribution images, \ie{}, the unknown unknowns.
Moreover, most learning-based methods for detecting image manipulation rely on the full supervision, \ie{}, training with pixel-level mask~\cite{salloum2018image,li2019localization,wu2019mantra,yang2020constrained,zhou2020generate,chen2021image,dong2022mvss}.
This approach is commonly adopted due to the creation of image manipulation datasets using sophisticated software~\cite{hsu2006detecting,dong2013casia,wen2016coverage,guan2019mfc}, where manipulated images and masks are generated simultaneously.
Although the pixel-mask can provide full supervision to help the model differentiate authentic and tampered image regions, the cost of such image annotation is non-trivial and limits the amount of the training images.
On the other hand, emerging language-driven image editing/synthesis or sketch-based manipulation methods do not necessarily generate pixel-level masks during the editing process~\cite{portenier2018faceshop,yu2019free,yang2020deep,zeng2021sketchedit,shi2021learning,jiang2021language,shi2021learning,patashnik2021styleclip}, but still have great potential to help train image manipulation detection if properly used.

To address the limitations of previous fully-supervised image manipulation detection methods, we propose weakly-supervised image
manipulation detection (W-IMD), where only binary image-level labels are required to tell whether a given image is authentic or tampered
with, thereby eliminating the need for pixel-level masks during training. 
We observe that image manipulation detection typically
relies on inconsistency detection between features of the manipulated regions compared to features from the surrounding regions.
Thus, we propose two different self-consistency learning schemes: (1) multi-source consistency (MSC) and (2) inter-patch consistency (IPC) to achieve weakly-supervised self-consistency learning (WSCL) that aims to
improve the generalization ability of image manipulation detection.

For (1) multi-source consistency (MSC) learning, we take advantage of
content-agnostic information by using different noise patterns in the
image~\cite{fridrich2012rich,bayar2018constrained} in a late-fusion manner.
Specifically, we build an exclusive model on different sources (\eg{}, raw RGB
image, and its noise maps) and generate an ensemble prediction by averaging
predictions from different models.
Intuitively, each source may focus on different locations, and locations where
all models have consistent high/low activations and are likely to be
manipulated/authentic.
Hence, we use the ensemble prediction as a pseudo ground truth to 
guide each individual model, and enable them to learn cross-source
information.
When combining predictions from different sources, the ensemble model can be more reliable and accurate than single models.
For (2) the inter-patch consistency (IPC) learning, it learns global pair-wise image patch-patch similarities in a
self-supervised learning manner.
By learning the pair-wise relationship, IPC helps the model to differentiate
low-level authentic and tampered image patch features.
Enforcing the IPC constraints helps to correct potential false positives, estimate a more complete image region of manipulation, and mitigate overfitting.

We conducted experiments on seven benchmark datasets to validate the effectiveness
of our weakly-supervised method.
First, we follow the conventional setting of image manipulation detection and
demonstrate that our \abbrmethod{} achieves comparable image-level detection
performances with several fully-supervised methods under both
in-distribution and out-of-distribution evaluations.
Furthermore, we validate that our method can be easily extended to new
manipulations where only image-level labels are available.
By fine-tuning on the image-level labels, our \abbrmethod{} achieves even better performance.
Finally, we also demonstrate that our method achieves reasonable pixel-level manipulation localization performance even under the setting of weakly-supervised learning.

To summarize, our contributions are threefold.
\begin{itemize}
    \item We first propose weakly-supervised image manipulation detection (W-IMD),
    where only binary image-level labels are required to achieve image
    manipulation detection. Such a paradigm eliminates the need for pixel-level
    annotations and can be easily adapted to new mask-free image editing
    techniques.
    \item We propose weakly-supervised self-consistency learning (WSCL) for the W-IMD task. By exploiting the multi-source consistency and inter-patch consistency, our \abbrmethod{} learns and fuses information from different content-agnostic sources, performs global image patch-patch relationship learning, and promotes generic image manipulation detection. Our WSCL also has the capability to locate the manipulation region in the pixel-level\footnote{In this paper, we use ``detection'' to indicate image-level fake/authentic classification, and use ``localization'' to indicate pixel-level
manipulation localization.}. 
    \item Experiments validate that our \abbrmethod{} achieves strong in-distribution and
    out-of-distribution image manipulation detection capability, promising results when fine-tuned with image-level labels on novel manipulations, and reasonable
    manipulation localization ability.
\end{itemize}

\section{Related Work}
\label{sec:related-work}

\begin{figure*}[t]
    \centering
    \includegraphics[width=\linewidth]{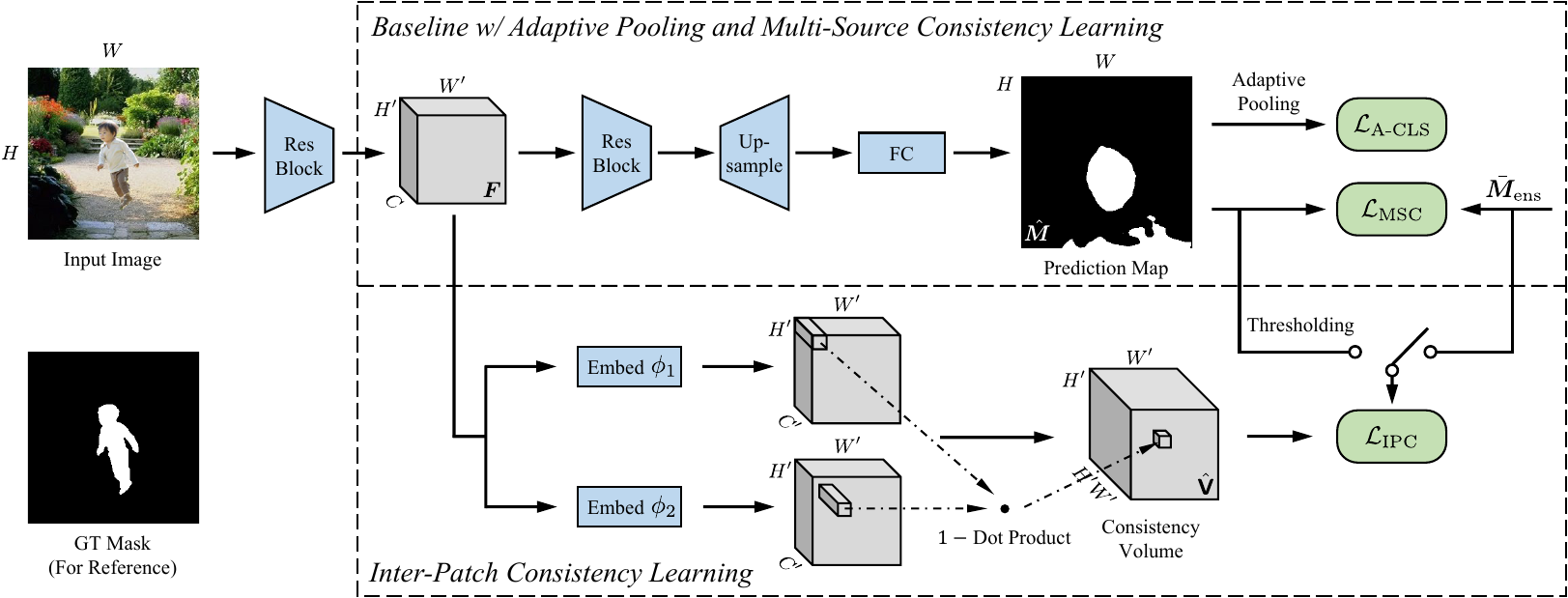}
    \caption{The single-stream overview. Given an input image, a baseline method
    (upper) predicts a manipulation map. The prediction map is supervised by an
    adaptive pooling-based classification loss $\mathcal{L}_{\text{A-CLS}}$ and
    a multi-source consistency learning loss $\mathcal{L}_{\text{MSC}}$.
    Meanwhile, the inter-patch consistency branch (bottom) learns a consistency
    volume measuring patch-patch similarities. The consistency volume is
    supervised by the inter-patch consistency loss $\mathcal{L}_{\text{IPC}}$.}
    \label{fig:framework}
\end{figure*}

\noindent\textbf{Image manipulation detection}.
We focus on detecting three types of image manipulation, \ie{},
copy-move~\cite{cozzolino2015efficient,rao2016deep,wen2016coverage,wu2018busternet,wu2018image},
splicing~\cite{cozzolino2015splicebuster,huh2018fighting,wu2017deep,salloum2018image},
and inpainting~\cite{zhu2018deep}.
Specifically, copy-move denotes copying and pasting image content within the
same image, and splicing indicates pasting image content from one image to
another.
Inpainting removes a particular area from an image and fills the region with new
content estimated from the surrounding.

Traditional unsupervised methods detect manipulations by exploiting specific
low-level tampering artifacts, such as color filter array (CFA)
analysis~\cite{ferrara2012image}, local noise analysis~\cite{mahdian2009using},
and double JPEG compression~\cite{bianchi2011improved}.
However, these methods assume that all given images consist of authentic and
tampered pixels, and perform two-class clustering on pixels to locate the
manipulation.
Thus, they detect manipulation out of all testing images, meaning that they
always achieve $0$ specificity and $1$ sensitivity.
Recent fully-supervised methods exploit content-agnostic features to localize
manipulations~\cite{zhou2018learning,li2019localization,wu2019mantra,hu2020span,yang2020constrained,chen2021image,dong2022mvss},
given the hypothesis that manipulation areas differ from pristine parts in terms
of their noise distributions.
Two of the noise filters are the most popular, \ie{}, the steganalysis rich
model (SRM) filter~\cite{fridrich2012rich,zhou2018learning} and the Bayar
convolutional filters~\cite{bayar2018constrained}.
Specifically, SRM filters~\cite{fridrich2012rich,zhou2018learning} use
predefined kernels to learn different types of noise residuals among the
neighboring pixels of the center pixel, followed by linear or non-linear max/min
operations.
The Bayar convolutional filters~\cite{bayar2018constrained} improve the SRM
filters by using learnable weights, with the constraint that the weighted sum of
neighboring pixels equals the negative of the weight of the center pixel.
In addition to the SRM and Bayar filters, CAT-Net~\cite{kwon2021cat} learns
compression artifacts from the RGB and DCT domains jointly.
MVSS-Net~\cite{chen2021image} learns semantic-agnostic information by exploiting
noise distribution and boundary artifacts in a multi-view, multi-scale fashion.
Except for leveraging noise maps, GSR-Net~\cite{zhou2020generate} designs a
pipeline to automatically generate copy-move images to enhance the training set.
Mantra-Net~\cite{wu2019mantra} learns the manipulation trace by conducting
fine-grained manipulation type classification.

There exist several works that exploit the image
consistency~\cite{bondi2016first,huh2018fighting,mayer2018learned,mayer2019forensic,mayer2020exposing,zhao2021learning}
for image forensics and Deepfake detection.
And most of them use a Siamese network to determine whether two input image
patches contain the same forensic characteristics, such as EXIF
metadata~\cite{huh2018fighting} or camera model
characteristics~\cite{bondi2016first,mayer2018learned,mayer2019forensic}.
Recently, Zhao~\etal{}~\cite{zhao2021learning} propose to use a pair-wise
similarity consistency volume~\cite{dosovitskiy2015flownet,zhao2018recognize} to
detect and localize Deepfakes in a fully-supervised setting.
Unlike Zhao~\etal{}~\cite{zhao2021learning} that requires a curated
inconsistency image generator and pixel-level ground truth to learn the
consistency volume, we only leverage image-level labels and use a
self-supervised approach for training.

\noindent\textbf{Weakly-supervised learning} aims to use coarse or incomplete
supervision to construct a model to predict fine-grained labels.
For example, given image-level categorical labels to predict bounding
box~\cite{kantorov2016contextlocnet,tang2018pcl} or segmentation
mask~\cite{pathak2014fully,pinheiro2015image,zhou2018weakly} and given
video-level categorical labels to predict temporal boundaries of
actions~\cite{wang2017untrimmednets}.
Such a paradigm greatly relieves the annotation burden from its fully-supervised
counterparts.

Our weakly-supervised image manipulation detection (W-IMD) is most related to
weakly-supervised semantic segmentation (W-SSS), where only image-level labels
can be leveraged to predict segmentation
mask~\cite{pathak2014fully,pinheiro2015image,zhou2018weakly,chan2021comprehensive}.
Different from most W-SSS works, this paper focuses on improving the
generalization ability of \emph{image-level} manipulation detection, instead of
pursuing high pixel-level localization ability.
Thus, we leverage two different single-stage W-SSS methods as baselines due to
their fast and simple training schemes: multiple instance learning fully
convolutional network (MIL-FCN)~\cite{pathak2014fully} and Araslanov and
Roth~\cite{araslanov2020single}.
The former applies multiple-instance learning on weakly-supervised segmentation
without considering specific prior knowledge on this task; the latter achieves
strong segmentation performance by considering several priors in W-SSS, such as
local consistency, semantic fidelity, and completeness.

\section{Proposed Method}
\label{sec:proposed-method}

During training, for each input image $\bm{X} \in \mathbb{R}^{ H\times W\times
3}$ with height $H$ and width $W$, we only have its image-level manipulation
label $y\in \{0, 1\}$, where $0$ denotes authentic images, and $1$ denotes
manipulated images.
During inference, for each image, we not only predict whether the image is
tampered with, but we also generate a binary localization map $\bar{\bm{M}}\in \{ 0, 1
\}^{H\times W}$ to localize manipulation at the pixel level.
An overview of our method is shown in \cref{fig:framework}.

\subsection{Baseline}
\label{subsec:baseline}

Without loss of generality, given an input image of size $H\times W$, we denote
the final prediction map generated by a baseline method as $\hat{\bm{M}} \in (0,
1)^{H\times W}$.
The image-level prediction $\hat{y}$ is generated by pooling the prediction map:
$\hat{y} = \text{Pool}(\hat{\bm{M}})$, where the pooling function can be
method-specific, \eg{}, global max pooling.
The image classification loss $\mathcal{L}_{\text{CLS}}$ is typically a binary
cross-entropy~(BCE) loss between the prediction and the ground truth
$\mathcal{L}_{\text{CLS}} = \mathcal{L}_{\text{BCE}} (y, \hat{y})$, where
\begin{equation}
    \mathcal{L}_{\text{BCE}}(y, \hat{y}) = -y \log (\hat{y}) - (1 -
y) \log (1 - \hat{y}).
    \label{eq:bce-loss}
\end{equation}
The final manipulation localization map $\bar{\bm{M}}$ is obtained by
thresholding $\hat{\bm{M}}$ at $\theta$, and $\theta$ is a thresholding
hyperparameter.

\subsection{Adaptive Pooling for Image-Level Detection}

Global max pooling has been one of the most widely used pooling methods for
image-level prediction generation in image manipulation
detection~\cite{salloum2018image,li2019localization,wu2019mantra,zhou2020generate,chen2021image}.
However, it has several clear disadvantages.
First, it only detects the most discriminative part, but it fails to detect the full
extent of the manipulation.
Second, the loss only back-propagates through the sole maximal response, impeding
the model training.
To address this problem, inspired by Otsu's method of image
binarization~\cite{otsu1979threshold}, we propose an adaptive pooling method,
which dynamically selects pixel-level responses from the prediction map
$\hat{\bm{M}}$ to form the image-level prediction $\hat{y}_{\text{A}}$.
 
Specifically, we first use Otsu's method to partition the prediction map into
two groups.
The Otsu's method finds a threshold $\omega_{\text{o}}$ that minimizes the
intra-class prediction variance~\cite{otsu1979threshold}:
\begin{equation}
    \begin{aligned}
        \omega_{\text{o}} = \mathop{\arg\min}\limits_{\omega \in \{
        \hat{m}_{i,j} \}}
        & \left| \{ \hat{m}_{i,j} | \hat{m}_{i,j} < \omega \} \right| \text{var}
        \left(\{ \hat{m}_{i,j} | \hat{m}_{i,j} < \omega \} \right) + \\
        & \left| \{ \hat{m}_{i,j} | \hat{m}_{i,j} \geq \omega \} \right|
        \text{var} \left(\{ \hat{m}_{i,j} | \hat{m}_{i,j} \geq \omega \}
        \right),
    \end{aligned}
\end{equation}
where $\text{var}(\cdot)$ denotes the variance, and $\hat{m}_{i,j}$ is the
pixel-level response at spatial location $(i,j)$ on $\hat{\bm{M}}$.
As Otsu's method only applies to discrete distributions, in practice, we
restrict the candidate set of thresholds $\omega$ to the values of pixel-level
responses $\{ \hat{m}_{i,j} \}$.
The image-level prediction $\hat{y}_{\text{A}}$ determined by our adaptive
pooling is the average value of the group with a higher response:
$\hat{y}_{\text{A}} = \frac{1}{|\mathbb{M}_{\text{h}}|} \sum_{\hat{m} \in
\mathbb{M}_{\text{h}}} \hat{m}$, where $\mathbb{M}_{\text{h}} = \{ \hat{m}_{i,
j} | \hat{m}_{i, j} \geq \omega_{\text{o}} \}$.
And our adaptive pooling-based classification loss $\mathcal{L}_{\text{A-CLS}}$
is as the BCE loss between the ground truth and the adaptive pooling prediction:
$\mathcal{L}_{\text{A-CLS}} = \mathcal{L}_{\text{BCE}} (y, \hat{y}_{\text{A}})$.
By exploiting multiple high responses instead of only the maximal one, our
adaptive pooling is more robust to noisy high responses and able to capture a
more complete manipulation region.

\subsection{Learning Multi-Source Consistency}

\begin{figure}
    \centering
    \includegraphics[width=\linewidth]{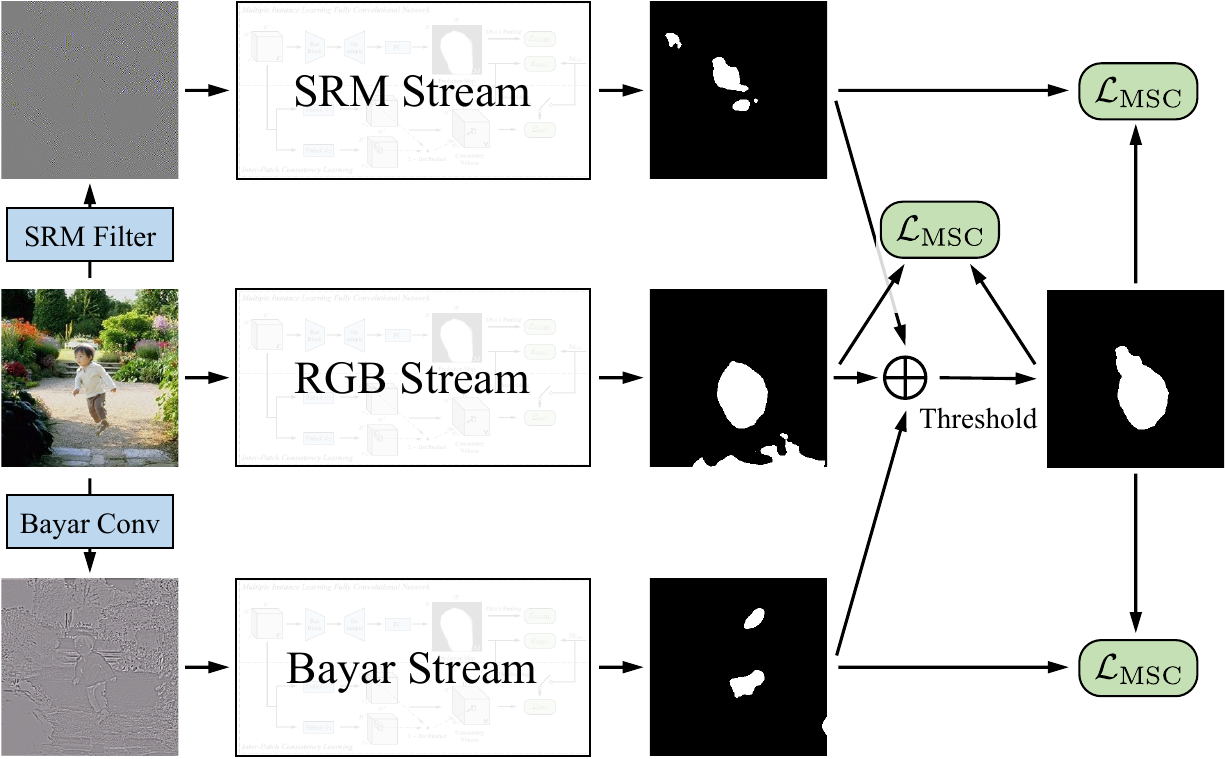}
    \caption{An illustration of multi-source consistency learning. Three
    parallel streams are trained on RGB image, SRM noise map, and Bayar noise
    map separately. Their weighted average prediction is used as pseudo ground
    truth to supervise each single stream.}
    \label{fig:multi-source-consistency}
\end{figure}

Prior arts found that exploring semantic information from images works well for
IND manipulation detection, but yields poor OOD detection
performance~\cite{zhou2020generate}.
Moreover, leveraging image noise maps to learn content-agnostic information can
produce strong
performance~\cite{fridrich2012rich,zhou2018learning,bayar2018constrained,wu2019mantra,hu2020span,chen2021image}.
Given these findings, we speculate that relying exclusively on content-related
information would be insufficient for detecting and localizing the
manipulations. 
However, the success of previous research heavily relies on the pixel-level
ground truth.
Without such strong supervision, the model easily overfits without fully
learning from different input sources.
To mitigate this problem, we propose multi-source consistency (MSC) learning.
First, MSC employs a multi-stream framework, with each stream exploiting
different sources of the image, and thus fully exploits the manipulation in
different views.
By combining the outputs of individual streams, the detection and localization
results can be more robust and accurate.
Moreover, the ensemble prediction is used to supervise each individual stream,
while helping to alleviate overfitting, correct failures within single streams,
and ultimately improve the final prediction.
An illustration is shown in~\cref{fig:multi-source-consistency}.

Specifically, we adopt a three-stream framework, with each stream taking as
input raw RGB image, SRM noise map~\cite{fridrich2012rich,zhou2018learning}, and
Bayar noise map~\cite{bayar2018constrained}, respectively.
The SRM and Bayar noise maps are selected due to its wide use in previous works~\cite{zhou2018learning,li2019localization,wu2019mantra,hu2020span,yang2020constrained,chen2021image,dong2022mvss}.
The three streams do not share parameters, but share the same training scheme.
Given predictions from three streams, an ensemble prediction
$\hat{\bm{M}}_{\text{ens}}$ can be obtained by weighted averaging three
prediction maps:
\begin{equation}
    \hat{\bm{M}}_{\text{ens}} = \frac{w_{\text{R}} \hat{\bm{M}}_{\text{R}} +
    w_{\text{S}} \hat{\bm{M}}_{\text{S}} + w_{\text{B}}
    \hat{\bm{M}}_{\text{B}}}{w_{\text{R}} + w_{\text{S}} + w_{\text{B}}},
\end{equation}
where $w_{\text{R}}$, $\hat{\bm{M}}_{\text{R}}$, $w_{\text{S}}$,
$\hat{\bm{M}}_{\text{S}}$, $w_{\text{B}}$, $\hat{\bm{M}}_{\text{B}}$ are
predefined weights and prediction maps of the RGB stream, SRM stream, and Bayar
stream, respectively.
Then, a binary pseudo ground truth $\bar{\bm{M}}_{\text{ens}}$ is generated by
thresholding the weighted average prediction map at $\theta$, and provides
pixel-level supervision of the three streams.

Our MSC learning encourages each individual stream to fit the pseudo ground
truth with a BCE loss:
\begin{equation}
    \mathcal{L}_{\text{MSC}} = \frac{1}{HW} \sum_{i,j}
    \mathcal{L}_{\text{BCE}}(\bar{m}_{\text{ens}, i,j}, \hat{m}_{\text{source},
    i, j}),
\end{equation}
where $\text{source} \in \{\text{R}, \text{S}, \text{B}\}$.
All individual streams are trained in parallel and simultaneously.
Intuitively, pixels that all three streams have high activations are more likely
to contain manipulations, while pixels that only one stream has high activations
are less likely to be manipulated.
Thus, our MSC enables each stream to improve itself by learning the voting
ensemble, and in turn improves the pseudo ground truth.

\subsection{Learning Inter-Patch Consistency}

Only exploiting local image features
fails when the tampered region is larger than the final layer's
receptive field, as tampered regions in manipulations like copy-move and
splicing are both from authentic images, and they both have unified, authentic
forensic characteristics.
Without referring to the global context, it is intractable to detect these
manipulations.
Thus, we propose to learn global inter-patch similarity (IPC) to detect
inconsistent image patches, and differentiate low-level features between
authentic and tampered patches.

IPC is conducted at an intermediate feature map $\bm{F} \in \mathbb{R}^{H'
\times W' \times C}$, where $H'$, $W'$, and $C$ are respectively height, width,
and the number of channels.
Each feature vector $\bm{f}_{i,j}$ at spatial location $i,j$ represents a local
image patch within its receptive field~\cite{long2015fully,zhao2021learning}.
For each patch, we compute its dot product similarity against all image
patches, thus, a consistency volume $\hat{\tens{V}} \in (0, 1)^{H' \times W' \times H'
\times W'}$ containing all pair-wise similarities can be obtained:
\begin{equation}
    \hat{v}_{i,j,h,k} = 1 - \sigma \left( \frac{\phi_{1} (\bm{f}_{i,j})\cdot
    \phi_{2} (\bm{f}_{h,k})}{\sqrt{C}} \right),
\end{equation}
where $\phi_{1}$ and $\phi_{2}$ are two embedding heads realized by two-layer
MLPs, $\hat{v}_{i,j,h,k}$ denotes the value at location $(i,j,h,k)$ of the
consistency volume, and $\sigma(\cdot)$ denotes the sigmoid function.
If patches $\bm{f}_{i,j}$ and
$\bm{f}_{h,k}$ share the same forensic characteristic (\ie{}, if they are both
authentic or both tampered with), then $\hat{v}_{i,j,h,k}=0$, while
$\hat{v}_{i,j,h,k}=1$ indicates different forensic characteristics.
Therefore, authentic images are expected to have all zero consistency volumes,
while tampered images should contain at least one location of value $1$ in their
consistency volumes.
\cref{fig:framework} bottom branch shows an illustration.

Despite the previous exploration where the consistency volume is trained with
full supervision and a hand-crafted inconsistency image
generator~\cite{zhao2021learning}, we show that the consistency volume can be
learned in a self-supervised learning way, and benefit from multi-source
consistency learning.

In the fully-supervised setting, such consistency volume can be easily learned
from the pixel-level ground truth~\cite{zhao2021learning}.
However, in the weakly-supervised setting, such ground truth is unavailable.
To mitigate this problem, we exploit two different ways to generate the learning
target of IPC.

\noindent\textbf{Self-supervision} is inspired by
self-distillation~\cite{zhang2021self}, where the intuition is that deeper
layers enjoy larger receptive fields and can make better predictions compared
to shallow layers.
In this setting, we regard the final localization map $\bar{\bm{M}}$ as the
teacher to guide the consistency volume learning within each individual stream.

\noindent\textbf{Ensemble-supervision} uses the MSC ensemble localization map
$\bar{\bm{M}}_{\text{ens}}$ as the IPC learning target.
Ensemble supervision fuses predictions from multiple sources and can help
individual streams learn cross-source information.

Given the target localization map $\bar{\bm{M}}_{\text{tgt}} \in \{
\bar{\bm{M}}, \bar{\bm{M}}_{\text{ens}} \}$, we first downsample it to size
$(H', W')$, then convert it to the target consistency volume
$\bar{\tens{V}}_{\text{tgt}}$:
if location $(i, j)$ and $(h, k)$ share the same value in the downsampled map,
then $\bar{v}_{\text{tgt}, i,j,h,k} = 0$, otherwise, $\bar{v}_{\text{tgt},
i,j,h,k} = 1$.
The IPC is supervised by the BCE loss:
\begin{equation}
    \mathcal{L}_{\text{IPC}} = \frac{1}{H'W'H'W'} \sum_{i,j,h,k}
    \mathcal{L}_{\text{BCE}} (\bar{v}_{\text{tgt}, i,j,h,k}, \hat{v}_{i,j,h,k}).
\end{equation}
In this way, our IPC enhances the low-level feature representation,
differentiates authentic and tampered image patches in shallow layers and in
turn boosts the final prediction.

\begin{table}[t]
    \centering
    \resizebox{\linewidth}{!}{%
    \begin{tabular}{c|c|>{\centering}p{0.9cm}>{\centering}p{0.9cm}|ccc}
        \hline 
        Split & Dataset & \#au & \#tp & \#cpmv & \#splc & \#inpaint\tabularnewline
        \hline 
        \multirow{3}{*}{Train} & CASIAv2~\cite{dong2013casia} & 7,491 & 5,063 & 3,235 & 1,828 & 0\tabularnewline
        \cline{2-7} \cline{3-7} \cline{4-7} \cline{5-7} \cline{6-7} \cline{7-7} 
         & GIER~\cite{shi2020benchmark} & 4,189 & 4,190 & \multicolumn{3}{c}{\rule[3pt]{1cm}{0.5pt} N/A \rule[3pt]{1cm}{0.5pt}}\tabularnewline
         & IEdit~\cite{tan2019expressing} & 2,255 & 2,255 & \multicolumn{3}{c}{\rule[3pt]{1cm}{0.5pt} N/A \rule[3pt]{1cm}{0.5pt}}\tabularnewline
        \hline 
        Val & IMD2020~\cite{novozamsky2020imd2020} & 2,010 & 2,010 & \multicolumn{3}{c}{\rule[3pt]{1cm}{0.5pt} N/A \rule[3pt]{1cm}{0.5pt}}\tabularnewline
        \hline 
        \multirow{6}{*}{Test} & CASIAv1~\cite{dong2010casia} & 800 & 920 & 459 & 461 & 0\tabularnewline
         & Columbia~\cite{hsu2006detecting} & 183 & 180 & 0 & 180 & 0\tabularnewline
         & Coverage~\cite{wen2016coverage} & 100 & 100 & 100 & 0 & 0\tabularnewline
         & NIST16~\cite{guan2019mfc} & 0 & 563 & 68 & 288 & 208\tabularnewline
        \cline{2-7} \cline{3-7} \cline{4-7} \cline{5-7} \cline{6-7} \cline{7-7} 
         & GIER~\cite{shi2020benchmark} & 452 & 618 & \multicolumn{3}{c}{\rule[3pt]{1cm}{0.5pt} N/A \rule[3pt]{1cm}{0.5pt}}\tabularnewline
         & IEdit~\cite{tan2019expressing} & 401 & 445 & \multicolumn{3}{c}{\rule[3pt]{1cm}{0.5pt} N/A \rule[3pt]{1cm}{0.5pt}}\tabularnewline
        \hline 
    \end{tabular}%
    }
    \caption{Dataset details. ``cpmv'', ``splc'' are abbreviations for copy-move
    and splicing, respectively.}
    \label{tab:dataset}
\end{table}

\begin{table*}[t]
    \centering
    \resizebox{\linewidth}{!}{%
\begin{tabular}{c|c|cccc|cccc|cccc|cccc|c|c}
\hline 
\multirow{2}{*}{} & \multirow{2}{*}{Method} & \multicolumn{4}{c|}{CASIAv1} & \multicolumn{4}{c|}{Columbia} & \multicolumn{4}{c|}{Coverage} & \multicolumn{4}{c|}{IMD2020} & \multicolumn{2}{c}{Avg}\tabularnewline
 &  & AUC & Spe. & Sen. & I-F1 & AUC & Spe. & Sen. & I-F1 & AUC & Spe. & Sen. & I-F1 & AUC & Spe. & Sen. & I-F1 & AUC & I-F1\tabularnewline
\hline 
\multirow{2}{*}{\begin{turn}{90}
Un.
\end{turn}} & NOI1~\cite{mahdian2009using} & 0.500 & 0.000 & 1.000 & 0.000 & 0.500 & 0.000 & 1.000 & 0.000 & 0.500 & 0.000 & 1.000 & 0.000 & 0.500 & 0.000 & 1.000 & 0.000 & 0.500 & 0.000\tabularnewline
 & CFA1~\cite{ferrara2012image} & 0.482 & 0.000 & 1.000 & 0.000 & 0.344 & 0.000 & 1.000 & 0.000 & 0.525 & 0.000 & 1.000 & 0.000 & 0.500 & 0.000 & 1.000 & 0.000 & 0.500 & 0.000\tabularnewline
\hline 
\cellcolor{fullcolor} & Mantra-Net~\cite{wu2019mantra} & 0.141 & 0.000 & 1.000 & 0.000 & 0.701 & 0.000 & 1.000 & 0.000 & 0.491 & 0.000 & 1.000 & 0.000 & 0.719 & 0.000 & 1.000 & 0.000 & 0.513 & 0.000\tabularnewline
\cellcolor{fullcolor}& CR-CNN~\cite{yang2020constrained} & 0.766 & 0.224 & 0.930 & 0.361 & 0.783 & 0.246 & 0.961 & 0.392 & 0.566 & 0.070 & 0.967 & 0.131 & 0.617 & 0.112 & 0.936 & 0.200 & 0.683 & 0.271\tabularnewline
\cellcolor{fullcolor}& GSR-Net~\cite{zhou2020generate} & 0.502 & 0.011 & 0.994 & 0.022 & 0.502 & 0.011 & 1.000 & 0.022 & 0.515 & 0.000 & 1.000 & 0.000 & 0.505 & 0.008 & 0.998 & 0.014 & 0.506 & 0.019\tabularnewline
\cellcolor{fullcolor}& CAT-Net~\cite{kwon2021cat} & 0.630 & 0.328 & 0.762 & 0.459 & 0.849 & 0.373 & 0.782 & 0.505 & 0.572 & 0.093 & 0.902 & 0.169 & 0.721 & 0.132 & 0.872 & 0.229 & 0.693 & 0.157\tabularnewline
\cellcolor{fullcolor}\multirow{-5}{*}{\begin{turn}{90}
    Full
\end{turn}}& FCN+DA~\cite{chen2021image} & \uline{0.796} & 0.844 & 0.717 & \textbf{0.775} & 0.762 & 0.322 & 0.950 & 0.481 & 0.541 & 0.100 & 0.900 & 0.180 & \textbf{0.746} & 0.100 & 0.981 & 0.182 & 0.711 & 0.404\tabularnewline
\hline 
\cellcolor{weakcolor}& MIL-FCN~\cite{pathak2014fully} & 0.647 & 0.538 & 0.569 & 0.553 & 0.807 & 0.220 & 0.732 & 0.338 & 0.542 & 0.062 & 0.793 & 0.115 & 0.578 & 0.116 & 0.886 & 0.205 & 0.644 & 0.303\tabularnewline
 \rowcolor{mygray}\cellcolor{weakcolor}& MIL-FCN~\cite{pathak2014fully} + \abbrmethod{} & \textbf{0.829} & 0.795 & 0.690 & \uline{0.738} & \textbf{0.920} & 0.519 & 0.983 & \textbf{0.680} & \uline{0.584} & 0.440 & 0.714 & \textbf{0.544} & \uline{0.733} & 0.221 & 0.966 & \textbf{0.360} & \textbf{0.766} & \textbf{0.580}\tabularnewline
 \cellcolor{weakcolor}& Araslanov and Roth~\cite{araslanov2020single} & 0.642 & 0.458 & 0.542 & 0.496 & 0.773 & 0.127 & 0.902 & 0.223 & 0.560 & 0.077 & 0.746 & 0.140 & 0.665 & 0.126 & 0.832 & 0.219 & 0.660 & 0.270\tabularnewline
 \rowcolor{mygray}\cellcolor{weakcolor}\multirow{-4}{*}{\begin{turn}{90}
Weak
\end{turn}}& Araslanov and Roth~\cite{araslanov2020single} + \abbrmethod{} & \uline{0.796} & 0.638 & 0.726 & 0.679 & \uline{0.917} & 0.324 & 0.948 & \uline{0.483} & \textbf{0.591} & 0.220 & 0.838 & \uline{0.348} & 0.701 & 0.193 & 0.872 & \uline{0.316} & \uline{0.751} & \uline{0.456}\tabularnewline
\hline 
\end{tabular}%
    }
    \caption{Comparison with unsupervised manipulation localization methods and
    fully-supervised methods on image-level manipulation detection. The best and
    the second best results are noted with \textbf{boldface} and
    \uline{underlined}, respectively. Note that the results of CAT-Net~\cite{kwon2021cat} is reproduced by training on CASIAv2 only.}
    \label{tab:comp-with-sota-image}
\end{table*}

\begin{table}[t]
        \centering
        \resizebox{\linewidth}{!}{%
        \begin{tabular}{c|c|cc|cc|cc}
                \hline 
                \multirow{2}{*}{} & \multirow{2}{*}{Method} & \multicolumn{2}{c|}{GIER~\cite{shi2020benchmark}} & \multicolumn{2}{c|}{IEdit~\cite{tan2019expressing}} & \multicolumn{2}{c}{Avg}\tabularnewline
                 &  & AUC  & F1  & AUC  & F1  & AUC & I-F1\tabularnewline
                \hline 
                \cellcolor{fullcolor} & CAT-Net~\cite{kwon2021cat} & 0.508 & 0.336 & 0.532 & 0.476 & 0.502 & 0.406\tabularnewline
                \cellcolor{fullcolor} & FCN+DA~\cite{chen2021image} & 0.507 & \underline{0.428} & 0.539 & 0.489 & 0.523 & \uline{0.458}\tabularnewline
                \cellcolor{fullcolor}\multirow{-3}{*}{\begin{turn}{90}
                Full
                \end{turn}}& MVSS-Net~\cite{chen2021image} & 0.510 & 0.325 & 0.537 & 0.522 & 0.523 & 0.423\tabularnewline
                \hline 
                \cellcolor{weakcolor} & MIL-FCN~\cite{pathak2014fully} + \abbrmethod{} & \underline{0.574} & 0.320 & \underline{0.563} & \underline{0.556} & \uline{0.568} & 0.438\tabularnewline
                \cellcolor{weakcolor}\multirow{-2}{*}{\begin{turn}{90}
                Weak
                \end{turn}}& MIL-FCN~\cite{pathak2014fully} + \abbrmethod{} w/ fine-tune & \textbf{0.621} & \textbf{0.533} & \textbf{0.617} & \textbf{0.602} & \textbf{0.619} & \textbf{0.568}\tabularnewline
                \hline 
        \end{tabular}%
        }
        \caption{Image-level manipulation detection performance comparison on
        novel image manipulation
        datasets~\cite{shi2020benchmark,tan2019expressing}.}
        \label{tab:novel-datasets}
\end{table}

\subsection{Optimization and Inference}

The overall loss $\mathcal{L}_{\text{total}}$ is a weighted sum of the adaptive
pooling-based image classification loss $\mathcal{L}_{\text{A-CLS}}$, the MSC
loss $\mathcal{L}_{\text{MSC}}$, and the IPC loss $\mathcal{L}_{\text{IPC}}$:
\begin{equation}
    \mathcal{L}_{\text{total}} = \mathcal{L}_{\text{A-CLS}} + w(t)
    \lambda_{\text{MSC}} \mathcal{L}_{\text{MSC}} + w(t) \lambda_{\text{IPC}}
    \mathcal{L}_{\text{IPC}},
\end{equation}
where $\lambda_{\text{MSC}}$ and $\lambda_{\text{IPC}}$ are weighting
hyperparameters.
$w(t) = \exp({-5(1-t/T)^2})$ is a time-dependent Gaussian warming-up function,
where $t$ is the current epoch, and $T$ is the maximal number of epochs.
Such a warming-up function prevents learning from unreliable pseudo ground truth
at early training stages~\cite{tarvainen2017mean,laine2016temporal}.

The final image-level prediction is obtained by weighted averaging the
predictions from three different streams, and the prediction map is the ensemble
localization map $\bar{\bm{M}}_{\text{ens}}$.

\section{Experiments}
\label{sec:experiments}

\noindent\textbf{Datasets}.
Without explicitly mentioning, we train our method on CASIAv2~\cite{dong2013casia} only.
For the in-distribution (IND) evaluation, we use the CASIAv1
dataset~\cite{dong2010casia}.
For the out-of-distribution (OOD) evaluation, we use three datasets:
Columbia~\cite{hsu2006detecting}, Coverage~\cite{wen2016coverage} and
IMD2020~\cite{novozamsky2020imd2020}.
We further follow the convention to use NIST16~\cite{guan2019mfc} for the
pixel-level manipulation localization evaluation.
We use the IMD2020 dataset~\cite{novozamsky2020imd2020} for validation and
hyperparameter tuning.
IMD2020 contains $2,010$ real-life manipulated images, and we randomly sample
$2,010$ images from the real image set as the authentic image set.
To demonstrate the capacity of our method for novel manipulations, we carry out
experiments on recent language-driven image editing datasets
GIER~\cite{shi2020benchmark} and IEdit~\cite{tan2019expressing}.
Both datasets consist of paired images before and after editing, and the
manipulations are not constrained by copy-move, splicing, and inpainting.
To avoid data leakage from the paired images in the two datasets, we only sample either an authentic image or an edited image from each pair to form the training
set.
Details on the datasets are listed in~\cref{tab:dataset}.

\noindent\textbf{Evaluation metrics}.
For image-level manipulation detection, we report specificity, sensitivity, and
their F1 score (I-F1).
The area under receiver operating characteristic (AUC) is also reported as a
threshold-agnostic metric for image-level detection.
For pixel-level manipulation localization, we follow previous
methods~\cite{zhou2018learning,salloum2018image,zhou2020generate,chen2021image}
to compute pixel-level precision, recall, and their F1 (P-F1) score on tempered
images.
The overall performance of image- and pixel-level manipulation
detection/localization is measured by the harmonic mean of pixel-level and
image-level F1 scores~\cite{chen2021image}, denoted as combined F1 (C-F1), and
is sensitive to the lower value of P-F1 and I-F1.
To ensure a fair comparison, a decision threshold of $0.5$ is employed for all methods when performing F1 computations.

\noindent\textbf{Implementation details}.
Our method is implemented with PyTorch~\cite{pazeke2019pytorch}.
We use ResNet50~\cite{he2016deep} as the backbone, and its weight is randomly
initialized.
The input size $H\times W$ is set to $224\times 224$.
Only random cropping and flipping are used for data augmentation.
We use the AdamW optimizer~\cite{loshchilov2018decoupled} with a learning rate
that decays from $10^{-4}$ to $10^{-5}$ and a weight decay factor $5\times
10^{-4}$.
We train the model for $60$ epochs, and the learning rate decays by a factor of
$0.1$ at the $50$-th epoch.
Following~\cite{chen2021image}, we use $\theta=0.5$ as the default threshold to
conduct the experiments.
The hyperparameters are determined via a grid search on the validation set:
$\lambda_\text{MSC}=0.1$, $\lambda_\text{IPC}=0.1$, and
$w_{\text{R}}:w_{\text{S}}:w_{\text{B}} = 1:2:2$.
The consistency volume is computed after the first residual block in
ResNet50, and its size is $H'\times W' = 56\times 56$.

For unsupervised methods, we use implementations provided by the
MKLabl\footnote{\url{https://github.com/MKLab-ITI/image-forensics}}, and block
size of $2$ are used for both CFA1~\cite{ferrara2012image} and
NOI1~\cite{mahdian2009using} algorithms.
For the results of fully-supervised methods (HP-FCN~\cite{li2019localization},
Mantra-Net~\cite{wu2019mantra}, CR-CNN~\cite{yang2020constrained}, and
GSR-Net~\cite{zhou2020generate}) on NIST16, Columbia, CASIAv1 and Coverage, we
use the reproduced results provided by~\cite{chen2021image,dong2022mvss}.
And the results of the rest of the data are reproduced by us.

\begin{table*}[t]
    \centering
    \resizebox{\linewidth}{!}{%
    \begin{tabular}{c|c|ccccc|c|cccc|c}
        \hline 
        \multirow{2}{*}{} & \multirow{2}{*}{Method} & \multicolumn{6}{c|}{Pixel-Level F1} & \multicolumn{5}{c}{Combined F1}\tabularnewline
         &  & CASIAv1 & Columbia & Coverage & IMD2020 & NIST16 & Avg & CASIAv1 & Columbia & Coverage & IMD2020 & Avg\tabularnewline
        \hline 
        \multirow{2}{*}{\begin{turn}{90}
        Un.
        \end{turn}} & NOI1~\cite{mahdian2009using} & 0.157 & 0.311 & 0.205 & 0.124 & 0.089 & 0.190 & 0.000 & 0.000 & 0.000 & 0.000 & 0.000\tabularnewline
         & CFA1~\cite{ferrara2012image} & 0.140 & 0.320 & 0.188 & 0.111 & 0.106 & 0.188 & 0.000 & 0.000 & 0.000 & 0.000 & 0.000\tabularnewline
        \hline 
        \cellcolor{fullcolor} & Mantra-Net~\cite{wu2019mantra} & 0.155 & 0.364 & 0.286 & 0.122 & 0.000 & 0.185 & 0.000 & 0.000 & 0.000 & 0.000 & 0.000\tabularnewline
        \cellcolor{fullcolor}& CR-CNN~\cite{yang2020constrained} & 0.405 & 0.436 & 0.291 & - & 0.238 & - & 0.382 & 0.413 & 0.181 & - & -\tabularnewline
        \cellcolor{fullcolor}& GSR-Net~\cite{zhou2020generate} & 0.387 & 0.613 & 0.285 & 0.175 & 0.283 & 0.349 & 0.042 & 0.042 & 0.000 & 0.026 & 0.028\tabularnewline
        \cellcolor{fullcolor}& CAT-Net~\cite{kwon2021cat} & 0.276 & 0.352 & 0.134 & 0.102 & 0.138 & 0.200 & 0.345 & 0.406 & 0.149 & 0.144 & 0.261\tabularnewline
        \cellcolor{fullcolor}\multirow{-5}{*}{\begin{turn}{90}
            Full
        \end{turn}}& FCN+DA~\cite{chen2021image} & 0.441 & 0.223 & 0.199 & 0.270 & 0.167 & 0.260 & 0.562 & 0.305 & 0.189 & 0.217 & 0.318\tabularnewline
        \hline 
        \cellcolor{weakcolor}& MIL-FCN~\cite{pathak2014fully} & 0.117 & 0.089 & 0.121 & 0.097 & 0.024 & 0.090 & 0.193 & 0.141 & 0.118 & 0.131 & 0.146\tabularnewline
         \rowcolor{mygray} \cellcolor{weakcolor} & MIL-FCN~\cite{pathak2014fully} + \abbrmethod{} & 0.172 & 0.270 & 0.178 & 0.193 & 0.110 & 0.185 & 0.280 & 0.386 & 0.268 & 0.252 & 0.296\tabularnewline
         \cellcolor{weakcolor}& Araslanov and Roth~\cite{araslanov2020single} & 0.112 & 0.102 & 0.127 & 0.094 & 0.026 & 0.092 & 0.182 & 0.140 & 0.133 & 0.046 & 0.125\tabularnewline
         \rowcolor{mygray} \cellcolor{weakcolor} \multirow{-4}{*}{\begin{turn}{90}
            Weak
        \end{turn}}& Araslanov and Roth~\cite{araslanov2020single} + \abbrmethod{} & 0.153 & 0.362 & 0.201 & 0.173 & 0.099 & 0.198 & 0.250 & 0.414 & 0.255 & 0.159 & 0.270\tabularnewline
        \hline 
    \end{tabular}%
    }
    \caption{Comparison with unsupervised and fully-supervised methods on
    pixel-level manipulation localization and the combined F1 score between I-F1
    and P-F1. The pixel-level manipulation localization performances are
    measured on manipulated images only.}
    \label{tab:comp-with-sota-pixel-and-comb}
\end{table*}

\subsection{Comparison with the State-of-the-art}
\label{subsec:comparison-with-sota}

As our essential goal is to improve the generalization ability for \emph{image-level}
manipulation detection, we build our \abbrmethod{} upon two single-stage W-SSS methods
for their fast and simple training schemes: MIL-FCN~\cite{pathak2014fully} and Araslanov and
Roth~\cite{araslanov2020single}.
The former is a multiple instance learning based method without considering
priors in semantic information; the latter is a state-of-the-art single-stage W-SSS method with several
domain-specific designs.

\noindent\textbf{Image-level manipulation detection} results are shown in
\cref{tab:comp-with-sota-image}.
Our \abbrmethod{} improves both MIL-FCN~\cite{pathak2014fully} and Araslanov and
Roth~\cite{araslanov2020single} baselines on all datasets.
Our \abbrmethod{} with both baselines compares favorably with the previous
fully-supervised methods in terms of detection AUC and F1.
We note that our \abbrmethod{} with Araslanov and Roth
baseline~\cite{araslanov2020single} underperforms that with the MIL-FCN
baseline~\cite{pathak2014fully}.
Such results indicate that the priors in W-SSS (\eg{}, local consistency and semantic
fidelity) may not help in W-IMD, and developing specific methods for W-IMD is
imperative.
We evaluate the IND and OOD manipulation detection with the MIL-FCN
baseline~\cite{pathak2014fully} in~\cref{fig:teaser}, where we observe a strong
OOD manipulation detection performance of our method that surpasses previous
fully-supervised methods, showing the effectiveness of our \abbrmethod{}.
Note that for unsupervised methods~\cite{mahdian2009using,ferrara2012image}, we use
the maximal response on the prediction map as its image-level prediction.
They tend to detect manipulations in all images, resulting in near $0.5$ AUC and
$0.0$ F1 scores.

\noindent\textbf{Novel manipulation detection}.
As emerging novice-friendly manipulation
methods~\cite{shi2021learning,jiang2021language,shi2021learning,portenier2018faceshop,yu2019free,yang2020deep,zeng2021sketchedit}
do not necessarily generate pixel-level masks during their editing process,
existing fully-supervised methods cannot make use of these weakly-labeled data.
We investigate the capacity of fully-supervised methods and our
weakly-supervised method on two additional
datasets~\cite{shi2020benchmark,tan2019expressing}, which contain manipulations
that are different from the standard setting (\ie{}, copy-move, splicing, and
inpainting).
The results are summarized in~\cref{tab:novel-datasets}, where MIL-FCN is used
as baseline~\cite{pathak2014fully} due to its strong image-level manipulation
detection performance observed in~\cref{tab:comp-with-sota-image}.
Without fine-tuning, our method already outperforms fully-supervised
counterparts at the average AUC on both datasets.
Such results demonstrate a strong generalization ability of our \abbrmethod{}.
We further fine-tune our model with image-level labels on the two datasets, and
achieve the best performance.
Though the comparison between our fine-tuned model and fully-supervised methods
may not be fair, they cannot be trained without pixel-level mask, demonstrating
the necessity of developing weakly-supervised methods.

\noindent\textbf{Pixel-level manipulation localization} results are listed in
the left part of~\cref{tab:comp-with-sota-pixel-and-comb}.
Our method achieves reasonable pixel-level manipulation localization
performance, and the average performance on five datasets is comparable with fully-supervised Mantra-Net~\cite{wu2019mantra} and CAT-Net~\cite{kwon2021cat}.
Such a strong performance demonstrates the capability of our pixel-level
manipulation localization.

\noindent\textbf{Overall detection and localization performance} is summarized
in the right part of~\cref{tab:comp-with-sota-pixel-and-comb}.
Our method achieves a similar average performance with CAT-Net~\cite{kwon2021cat}.
Surprisingly, our method achieves the best overall performance on the Coverage dataset~\cite{wen2016coverage}.
Such a strong performance demonstrates the effectiveness of our method.

\begin{figure*}[!htb]
    \centering
    \includegraphics[width=\linewidth]{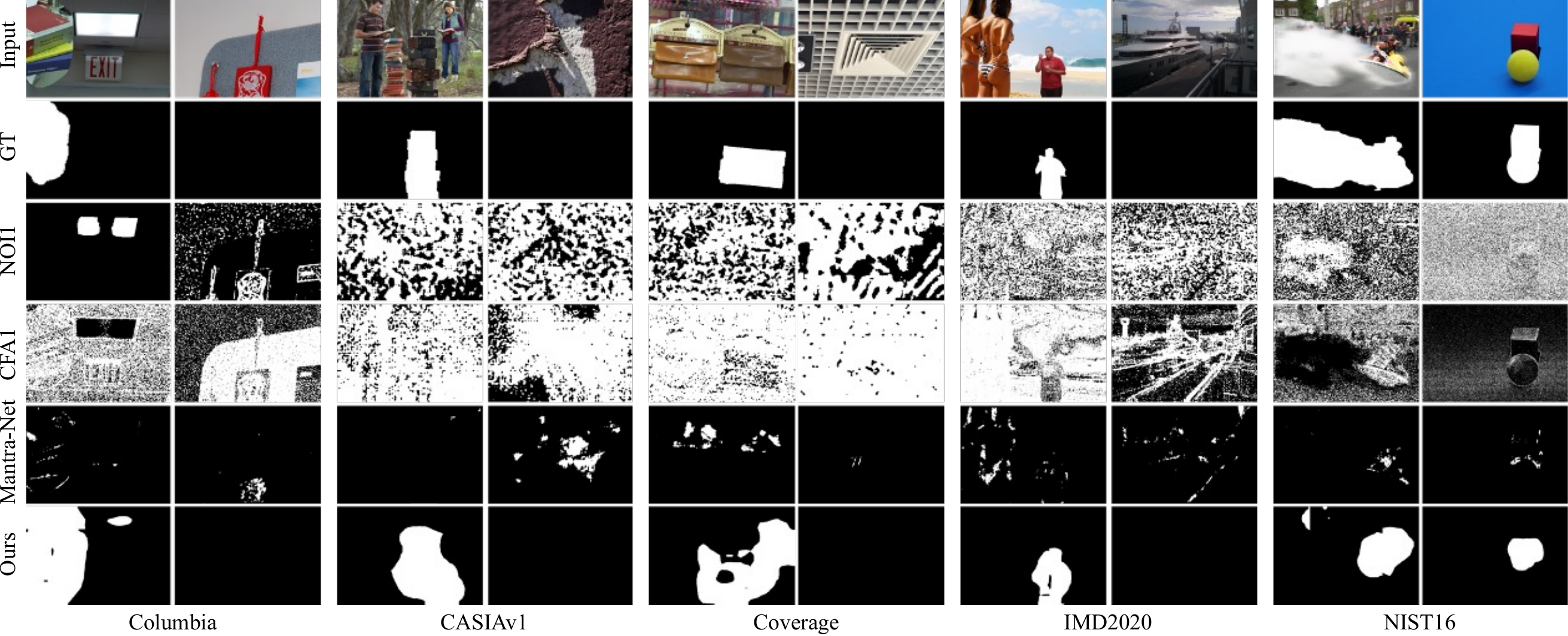}
    \caption{Qualitative results on five datasets. From top to bottom in each
    group: input image, ground truth mask, and predictions from
    NOI1~\cite{mahdian2009using}, CFA1~\cite{ferrara2012image},
    Mantra-Net~\cite{wu2019mantra} and our \abbrmethod{} with
    MIL-FCN~\cite{pathak2014fully} as the baseline.}
    \label{fig:qualitative-results}
\end{figure*}

\noindent\textbf{Qualitative results} are visualized
in~\cref{fig:qualitative-results}.
We make the following observations.
(1) Predictions from unsupervised
methods~\cite{mahdian2009using,ferrara2012image} tend to be noisy, while both
fully-supervised method and our weakly-supervised method generate clean
localization maps.
(2) Our method tends to predict a large area of manipulation, encompassing the
ground truth area, while fully-supervised method detects clean manipulation
boundaries.
(3) Our method shows degraded results on copy-move (see Coverage examples),
where both source and target manipulated areas are from the same image, and our
method tends to detect both areas.

\begin{table}[t]
    \centering
    \resizebox{0.9\linewidth}{!}{%
    \begin{tabular}{c|cccc|c|c}
        \hline 
        \multirow{2}{*}{Method} & \multicolumn{4}{c|}{Image-Level} & \multirow{2}{*}{P-F1} & \multirow{2}{*}{C-F1}\tabularnewline
         & AUC & Spe. & Sen. & I-F1 &  & \tabularnewline
        \hline 
        Max Pool & 0.578 & 0.116 & 0.886 & 0.205 & \textbf{0.131} & 0.131\tabularnewline
        \hline 
        Avg Pool & 0.569 & 0.076 & 0.902 & 0.140 & 0.082 & 0.103\tabularnewline
        \hline 
        GeM~\cite{radenovic2018fine} & \uline{0.683} & 0.139 & 0.725 & \uline{0.233} & 0.111 & 0.149\tabularnewline
        \hline 
        GSM~\cite{wang2017learning} & 0.679 & 0.105 & 0.833 & 0.186 & \uline{0.127} & \uline{0.151}\tabularnewline
        \hline 
        AP (Ours) & \textbf{0.693} & 0.162 & 0.788 & \textbf{0.269} & 0.116 & \textbf{0.162}\tabularnewline
        \hline 
    \end{tabular}%
    }
    \caption{Comparison on different pooling methods on the
    IMD2020~\cite{novozamsky2020imd2020} dataset with RGB as the source.}
    \label{tab:pooling}
\end{table}

\begin{table}[t]
    \centering
    \resizebox{0.85\linewidth}{!}{%
    \begin{tabular}{cc|cccc|c|c}
        \hline 
        \multirow{2}{*}{} & \multirow{2}{*}{Source} & \multicolumn{4}{c|}{Image-Level} & \multirow{2}{*}{P-F1} & \multirow{2}{*}{C-F1}\tabularnewline
         &  & AUC & Spe. & Sen. & F1 &  & \tabularnewline
        \hline 
        \hline 
        \multirow{4}{*}{\begin{turn}{90}
        w/o MSC
        \end{turn}} & RGB & 0.693 & 0.162 & 0.788 & 0.269 & 0.116 & 0.162\tabularnewline
         & Bayar & 0.685 & 0.187 & 0.642 & 0.290 & 0.132 & 0.181\tabularnewline
         & SRM & 0.674 & 0.196 & 0.733 & 0.309 & 0.109 & 0.161\tabularnewline
        \cline{2-8} \cline{3-8} \cline{4-8} \cline{5-8} \cline{6-8} \cline{7-8} \cline{8-8} 
         & Fusion & 0.701 & 0.209 & 0.748 & 0.327 & 0.143 & 0.199\tabularnewline
        \hline 
        \multirow{4}{*}{\begin{turn}{90}
        w/ MSC
        \end{turn}} & RGB & 0.701 & 0.185 & 0.836 & 0.303 & 0.136 & 0.188\tabularnewline
         & Bayar & \uline{0.715} & 0.193 & 0.762 & 0.308 & 0.177 & 0.225\tabularnewline
         & SRM & 0.707 & 0.210 & 0.837 & \uline{0.336} & \uline{0.181} & \uline{0.235}\tabularnewline
        \cline{2-8} \cline{3-8} \cline{4-8} \cline{5-8} \cline{6-8} \cline{7-8} \cline{8-8} 
         & Fusion & \textbf{0.726} & 0.218 & 0.857 & \textbf{0.348} & \textbf{0.188} & \textbf{0.244}\tabularnewline
        \hline 
    \end{tabular}%
    }
    \caption{Ablation study on the multi-source consistency learning on IMD2020~\cite{novozamsky2020imd2020}.}
    \label{tab:ablation}
\end{table}

\begin{table}[t]
    \centering
    \resizebox{0.75\linewidth}{!}{%
    \begin{tabular}{c|cccc|c|c}
        \hline 
        \multirow{2}{*}{IPC} & \multicolumn{4}{c|}{Image-Level} & \multirow{2}{*}{P-F1} & \multirow{2}{*}{C-F1}\tabularnewline
         & AUC & Spe. & Sen. & F1 &  & \tabularnewline
        \hline 
        \hline 
        w/o & 0.726 & 0.218 & 0.857 & 0.348 & 0.188 & 0.244\tabularnewline
        self. & \uline{0.730} & 0.219 & 0.920 & \uline{0.354} & \uline{0.192} & \uline{0.249}\tabularnewline
        ens. & \textbf{0.733} & 0.221 & 0.966 & \textbf{0.360} & \textbf{0.193} & \textbf{0.252}\tabularnewline
        \hline 
    \end{tabular}%
    }
    \caption{Ablation study on the inter-patch consistency learning on IMD2020~\cite{novozamsky2020imd2020}.}
    \label{tab:ablation}
\end{table}

\subsection{Ablation Study}
\label{subsec:ablation-study}

We carry out the ablation study on an OOD dataset
IMD2020~\cite{novozamsky2020imd2020} with MIL-FCN~\cite{pathak2014fully} as the
baseline, where all variants are trained with the CASIAv2
dataset~\cite{dong2013casia}.
The ablation study is performed progressively, with each subsequent setting using the previous one as a baseline.

\noindent\textbf{Adaptive pooling}.
To mitigate the problem of max pooling in previous methods, we propose an
adaptive pooling to dynamically assign image-level labels to the pixels.
We compare our adaptive pooling with related pooling
methods~\cite{wang2017learning,radenovic2018fine} in~\cref{tab:pooling}.
The results show adaptive pooling achieves the best performance on all major metrics.
Besides, both GSM and GeM introduce an additional hyperparameter, while our
adaptive pooling does not require any hyperparameter, making it more flexible.
Such advantage demonstrates the effectiveness of our adaptive pooling.

\noindent\textbf{Multi-source consistency learning} promotes unanimous
predictions among all individual models through a pixel-level pseudo ground
truth.
The results are summarized in~\cref{tab:ablation}.
We observe that the late fusion improves the single-stream performance w/ and w/o MSC,
showing the effect of voting ensemble.
Furthermore, our MSC improves the performance on single streams and the fusion
results, demonstrating its effectiveness on improving generalization.

\noindent\textbf{Inter-patch consistency learning} aims to learn global
patch-patch similarities, and further differentiate low-level authentic and
tampered image patch features.
Two different supervision implementations are tested: the localization map from
the same stream (self-supervision), and the ensemble localization map from three
streams (ensemble-supervision).
We make the following observations from~\cref{tab:ablation}.
(1) Both implementations of IPC clearly improve overall performance, showing the effectiveness of IPC.
(2) The ensemble-supervision IPC outperforms the self-supervision counterpart,
and this is intuitive as the ensemble target fuses information from multiple
sources.

\begin{figure}[t]
    \centering
    \subfloat[IND manipulation detection under JPEG compression.\label{subfig:ind-jpeg}]{\includegraphics[width=0.475\linewidth]{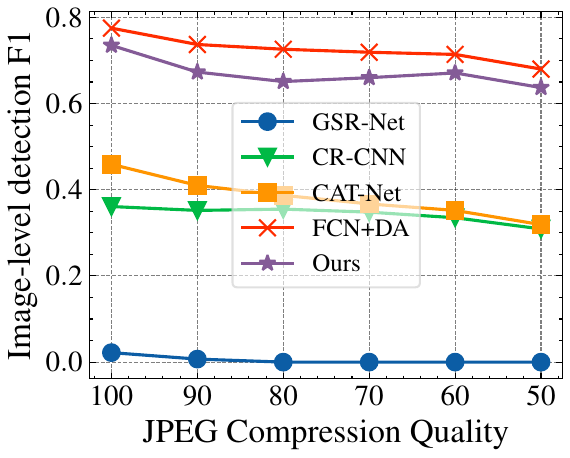}}
    \hfill
    \subfloat[OOD manipulation detection under JPEG compression.\label{subfig:ood-jpeg}]{\includegraphics[width=0.475\linewidth]{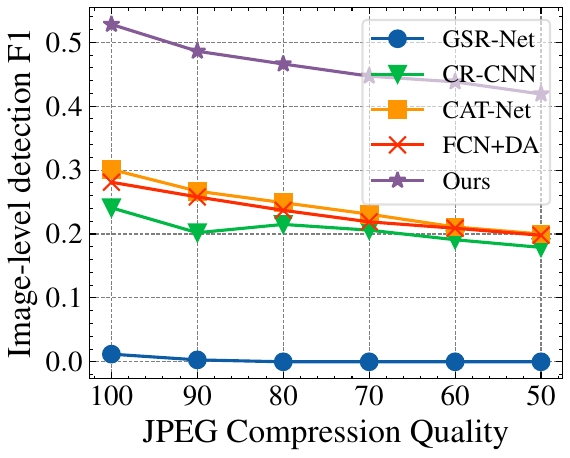}}

    \subfloat[IND manipulation detection under Gaussian blur.\label{subfig:ind-blur}]{\includegraphics[width=0.475\linewidth]{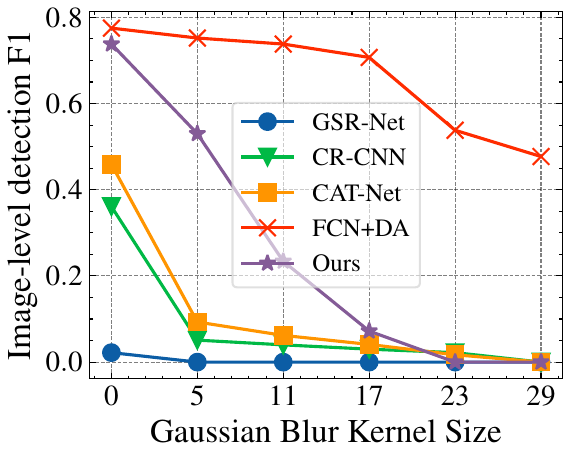}}
    \hfill
    \subfloat[OOD manipulation detection under Gaussian blur.\label{subfig:ood-blur}]{\includegraphics[width=0.475\linewidth]{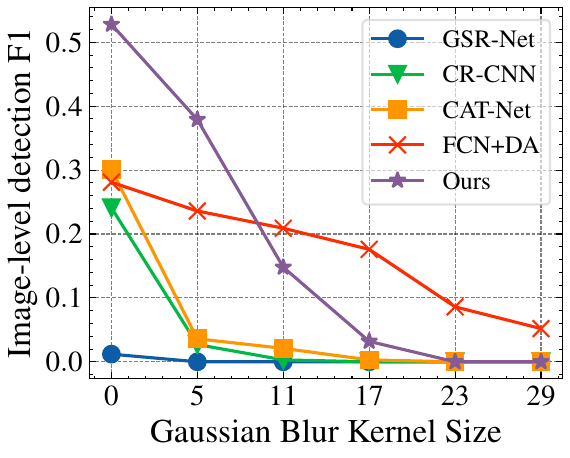}}

    \caption{Robustness evaluation against JPEG compression and Gaussian blur.
    All methods are trained on CASIAv1~\cite{dong2010casia}.
    CASIAv2~\cite{dong2010casia,dong2013casia} is used for IND testing, and the
    average results on Columbia~\cite{hsu2006detecting},
    Coverage~\cite{wen2016coverage} and IMD2020~\cite{novozamsky2020imd2020} are
    used for OOD testing. Our method is robust against JPEG compression, and
    mild Gaussian blur.}
    \label{fig:robustness}
 \end{figure}

\noindent\textbf{Robustness comparison with fully-supervised methods}.
We apply JPEG compression and Gaussian blur separately on
CASIAv1~\cite{dong2010casia} to evaluate the robustness of our method.
As shown in~\cref{subfig:ind-jpeg} and \cref{subfig:ood-jpeg}, our method
effectively defends against JPEG compression, especially under the OOD
evaluation, where our method significantly outperforms all competing methods.
As for the Gaussian blur, our method resists mild Gaussian blur, but is
vulnerable to the blur with large kernel sizes, as shown
in~\cref{subfig:ind-blur} and \cref{subfig:ood-blur}.
While this limitation highlights the need for further research and optimization
to improve our method's resistance to Gaussian blur, it also offers valuable
insights into the challenges of robustness in image manipulation detection.

\noindent\textbf{Ablation study on the early fusion architecture}.
In our method, we use a late fusion architecture to fuse multi-source
information.
We further evaluate our method on the early fusion architecture, where different
sources are concatenated in the channel dimension at input.
The results are listed in~\cref{tab:early-fusion}.
Note that multi-source consistency learning and ensemble-supervision inter-patch
consistency learning do not apply to the early fusion architecture, and thus are
excluded.
The results show both adaptive pooling and self-supervision inter-patch
consistency learning individually improve the performance of the early fusion
architecture, and their combination leads to the best performance, demonstrating
the effectiveness of our method under the early fusion architecture.
Furthermore, under most settings, the performances in early fusion underperform
their counterparts in late fusion.
Especially, early fusion performances even underperform several single stream
performances in late fusion.
Such results show early fusion architecture cannot fully utilize each single
source under the weakly-supervised setting, and a late fusion design is needed
for the weakly-supervised image manipulation detection and localization.

\begin{table}[t]
    \centering
    \resizebox{0.85\linewidth}{!}{%
    \begin{tabular}{cc|cccc|c|l}
        \hline 
        \multirow{2}{*}{AP} & \multirow{2}{*}{IPC} & \multicolumn{4}{c|}{Image-Level} & \multirow{2}{*}{P-F1} & \multirow{2}{*}{C-F1}\tabularnewline
         &  & AUC & Spe. & Sen. & F1 &  & \tabularnewline
        \hline 
        \hline 
        - & - & 0.597 & 0.134 & 0.700 & 0.225 & 0.126 & 0.161\tabularnewline
        \hline 
        \checkmark & - & 0.611 & 0.158 & 0.750 & 0.261 & 0.142 & 0.184\tabularnewline
        - & self & \uline{0.641} & 0.166 & 0.733 & \uline{0.271} & \uline{0.158} & \uline{0.200}\tabularnewline
        \hline 
        \checkmark & self & \textbf{0.682} & 0.219 & 0.771 & \textbf{0.341} & \textbf{0.177} & \textbf{0.233}\tabularnewline
        \hline 
    \end{tabular}%
    }
    \caption{Ablation study on early fusion architecture on
    IMD2020~\cite{novozamsky2020imd2020}, where the concatenation of RGB image,
    Bayar noise map and SRM noise map is fed into a single model. AP is an
    abbreviation for adaptive pooling.}
    \label{tab:early-fusion}
\end{table}

\section{Conclusion}
\label{sec:conclusion}

We propose the task of weakly-supervised image manipulation detection, such that
only binary image-level labels are required to detect and localize
manipulations.
We propose a weakly-supervised self-consistency learning for this task that aims
to improve the generalization ability.
Two different self-consistency learning schemes are employed: multi-source consistency and inter-patch consistency.
By leveraging content-agnostic information and combining predictions from various sources, MSC enhances the individual stream's ability for both manipulation detection and localization.
IPC learns the global similarity between image patches to detect a complete region of manipulation, which improves the low-level representation of image patches and thus facilitates the MSC learning process.
Our \abbrmethod{} shows strong image-level manipulation detection performance under
both IND and OOD evaluation settings.
We also achieve reasonable pixel-level manipulation localization performance.

\noindent\textbf{Limitations}.
While our method demonstrates robust performance in detecting image-level
manipulation, its capability in localizing the pixel-level manipulation is
merely satisfactory.
This is an important limitation as accurate localization is key to
explainability and understanding the extent of the manipulation,
which is vital in forensics applications.
Besides, as shown in our supplementary material, our method is vulnerability to
certain types of noise and distortions, such as Gaussian blur, which could
potentially be exploited to bypass our detection method.
This highlights the need for the method to be robust not just against various
manipulation techniques, but also against different types of image noise and
distortions.
Thus, future work should aim to improve the pixel-level manipulation localization
ability, and the robustness against different image distortions.

\section*{Acknowledgements}

This work is supported in part by the Defense Advanced Research Projects Agency (DARPA) under Contract No.~HR001120C0124. Any opinions, findings and conclusions or recommendations expressed in this material are those of the author(s) and do not necessarily reflect the views of the Defense Advanced Research Projects Agency (DARPA).

{\small
\bibliographystyle{ieee_fullname}
\bibliography{egbib}
}

\end{document}